# Attacks and Faults Injection in Self-Driving Agents on the Carla Simulator – *Experience Report*

Niccolò Piazzesi, Massimo Hong, Andrea Ceccarelli

Università degli Studi di Firenze, Dept. of Mathematics and Informatics, Florence, Italy

**Abstract.** Machine Learning applications are acknowledged at the foundation of autonomous driving, because they are the enabling technology for most driving tasks. However, the inclusion of trained agents in automotive systems exposes the vehicle to novel attacks and faults, that can result in safety threats to the driving tasks. In this paper we report our experimental campaign on the injection of adversarial attacks and software faults in a self-driving agent running in a driving simulator. We show that adversarial attacks and faults injected in the trained agent can lead to erroneous decisions and severely jeopardize safety. The paper shows a feasible and easily-reproducible approach based on open source simulator and tools, and the results clearly motivate the need of both protective measures and extensive testing campaigns.

## 1 Introduction

Machine Learning (ML) is at the foundation of the most relevant autonomous driving applications, with more and more usage in safety critical functionalities as for example obstacle detection, lane detection, traffic sign recognition, and ultimately self-driving. This requires that the ML-based applications (and the supporting hardware) are both safe and secure i.e., faults and attacks must not compromise system safety. However, the introduction of ML exposes novel attack surfaces as well as new possible fault modes.

Several works in the most recent years have investigated the effects of attacks and faults on ML-applications and have provided evidence that even small perturbations caused by hardware or software faults can deceive the ML-based application, to the extent that an incorrect output is produced [25], [24], [29]. While it is possible that a single transient fault does not lead to wrong decisions or it is masked through the activation of the different layers of the neural network [24], there is still the risk that residual software faults (when not hardware faults) manifest into an observable output corruption [26], [27], [29].

Additionally, recent works on adversarial machine learning [12] have shown that ML applications can be confused by malicious perturbations of the input i.e., by crafting inputs that are syntactically correct, but artificially modified such that the ML application is mistaken. Despite these adversarial attack are typically though for object classification, they could be applied as well to the inputs of a driving tasks [7], as we are also considering in this paper.

Consequently, solutions are required to evaluate the safety of ML applications against software and hardware faults, and to secure the system with respect to the new attack surfaces introduced by the ML applications themselves. It is not surprising that a proliferation of tools and supports to test ML-applications, including security and robustness testing, has been witnessed in recent years, altough with a general preference for classification and detection tasks [29].

Without aiming to develop new tools, but reusing existing ones, we experiment and discuss on the risk of adversarial attacks and software faults for an end2end self-driving agent, which directly maps sensory information to driving commands without organizing the self-driving pipeline in separate tasks (for example, it does not distinguish tasks in obstacle detection, localization, trajectory planning, etc.) [20]. To the best of our knowledge, no works experimented with end2end self-driving agents including both faults and attacks perspectives. Consequently, this experience report provides a guide as well as practical evidence of a method easy-to-implement that allows to rapidly deploy tests for self-driving agents under various adversarial attacks and faulty conditions, in an entirely simulated (and reproducible) environment. More in details, the paper explains how to configure, inject and ultimately collect evidences of the effects of attacks and faults injected. We show how to manipulate a driving simulator to perform the experiments and to execute the experimental campaign. Results give evidence of the effects of attacks and faults injection, especially measured in terms of wrong decisions of the trained agent that lead to collisions or traffic offences; we show that the trained agent fails under multiple injection conditions. The settings and source code we used are available at [6].

The rest of the paper is organized as follows. Section 2 reviews background notions on techniques and the supporting tools that will be used in the paper. Section 3 and Section 4 describe respectively how we performed adversarial attacks and fault injection in the self-driving agent. Section 5 presents the experimental campaign and results. Section 6 discusses limitations of the approach and it contextualizes our work with respect to the current scenarios of testing self-driving agents against faults and attacks. Finally, Section 7 concludes the paper.

## 2   Background

### 2.1   Adversarial Attacks using the ART Toolbox

Machine Learning models are vulnerable to adversarial examples, which are inputs (images, texts, tabular data, etc.) deliberately modified, while being perceptually indistinguishable, to produce a desired response by the model [17]. By adding small perturbations to original images, adversarial attacks can deceive a target model to produce completely wrong predictions [7]. In general, adversarial attacks are organized in three categories: evasion, poisoning, and extraction attacks, that we review below.

*Evasion attacks* modify the input to a classifier such that it is misclassified, while keeping the modification as small as possible. Evasion attacks can be black-box or white-box: in the white-box case, the attacker has full access to the architecture and parameters of the classifier. For a black-box attack, clearly this is not the case.

In *poisoning attacks*, attackers have the opportunity of manipulating the training data to significantly decrease the overall performance, cause targeted misclassification or bad behavior, and insert backdoors and neural trojans.

Last, *extraction attacks* aim to develop a new model, starting from a proprietary black-box model, that emulate the behavior of the original model.

The Adversarial Robustness Toolbox (ART, [8]) is a Python library originally developed by IBM, and recently donated to the Linux Foundation. It provides the tools to craft adversarial attacks (as well as to build defenses against them). A large set of attacks from the state of the art are implemented in ART, and they can be invoked just providing as input the classifier and other parameters as the loss and optimization functions and the size of the input images. ART supports the most known Machine Learning libraries as PyTorch and TensorFlow, and it is released with the MIT open source license.

### 2.2 Fault injection in trained agents with PytorchFI

As software fault model, we consider any software fault whose effect is modifying the value of weights or neurons in convolutional operations of the neural network during its execution. In this paper we use the tool PyTorchFi [24], [31] to modify the neuron or weights of the neural network, such that we can observe the consequences that the perturbations bring to the vehicle behaviour.

PyTorchFi is a runtime perturbation tool for deep neural networks, implemented for the PyTorch deep learning platform. It enables users to perform perturbation on weight or neurons of DNNs at runtime. PyTorchFi provides an easy-to-use API and an extensible interface, enabling users to choose from various perturbation models (or design their own custom models) [24].

PyTorchFi offers different default perturbation models that a user can select. In general, the steps to use its API are as follows: i) choose the error model, ii) specify the injection location; there can be either a single or multiple locations to perform multiple perturbations across the neural network (injection locations are specified by the layer, feature map, and neuron's coordinate position in the tensor); iii) specify whether to have the same perturbation across all elements in a batch, or a different perturbation per element; iv) perform the injection [24].

The injection occurs at runtime, by taking in the location of the erroneous neuron/weight and appending it to a list of positions in the tensor to change. Then, on every layer, the forward hook iterates through all of the locations and corrupt the corresponding value based on the selected perturbation model [24].

### 2.3 Carla Simulator and Learning by Cheating (LbC)

The Open Urban Driving Simulator Carla (Car Learning to Act, [1]) is a simulator for autonomous driving that has been implemented as an open-source layer over the Unreal

Engine 4 (UE4, [2]). Its aim is to support training, prototyping, and validation of autonomous driving models, including both perception and control. Carla includes urban layouts, several vehicle models, buildings, pedestrians, street signs, etc. The simulation platform supports flexible setup of sensor suites, and in particular we will exploit the camera sensor, that allows acquiring images from the frontal camera of the vehicle at a specified Frame Per Second (FPS) rate. Carla provides information on the simulated vehicles as position, orientation, speed, acceleration, collisions and traffic violations. Further, weather conditions and time of day can be specified.

Amongst the various ML-based applications that exist for Carla, in our work we prefer a self-driving agent over other agents e.g., object recognition agents. In fact, a self-driving agent allows showing the effect of persistent faults or continuous attacks applied on consequential images, rather than on individual images without a continuous context [4]. Amongst self-driving agents, we select the trained agent Learning by Cheating (LbC) developed by Chen et al. [3]. LbC is an end2end learning [20] approach for self-driving, which directly map sensory information to steering commands. The LbC model is organized in a ResNet-34 backbone pretrained on ImageNet and three up-convolutional layers. LbC has demonstrated very good performance, with a minimal number of collisions under most of the environmental and traffic conditions [3].

The main control loop of LbC is explained in Listing 1 [30]. Using the camera as its unique input sensor, at each simulation step it is acquired (through *env.get_observations* in Listing 1): i) one *RGB image* from the frontal camera of the vehicle at a resolution of $384 \times 160$ pixels, and ii) the current *speed* from the speed sensor. These values are processed by the trained agent (*agent.run_step* in Listing 1), together with an high-level *command* ("follow lane", "turn left", "turn right", "go straight") that describes the planned route. In this way, the trained agent predicts waypoints in the camera coordinates, and then, these waypoints are projected into the vehicle's coordinate image (essentially, a trajectory is designed). From this, a low-level controller is executed (*env.apply_control* in Listing 1), that decides the steering angle, the throttle level, and the braking force. Finally, throttle, speed and braking are applied on the vehicle [3]. Additional details on the LbC trained agent are outside the scope of this paper and are in [3], [30].

```
while env.tick(): #at each simulation step
    observations = env.get_observations() #collect RGB image and current speed
    control= agent.run_step(observations) #compute throttle, steer, brake
    diagnostic = env.apply_control(control) #apply computed throttle, steer, brake
```

**Listing 1.** The main decision loop in LbC [3], [30].

## 3 Injection of Attacks in a self-driving agent

We describe how we inject adversarial attacks in the self-driving agent LbC.

### 3.1 Selection of suitable attacks

We are relying only on evasion attacks in this work. In fact, poisoning attacks operate during the training phase of an agent: this type of attacks was not relevant to our objectives, because we are interested in investigating a trained model that runs on a vehicle. Extraction attacks are instead discarded as we are interested in safety conditions, and not on industrial secrecy thefts. Instead, evasion attacks have the likelihood to be carried out on a self-driving agent while it is executing, thus compromising safety.

We select 4 attacks that we think are interesting to inject: Spatial Transformation [10], HopSkipJump [11], Basic Iterative Method [12] and NewtonFool [13]. The four attacks are chosen because they represent different approaches to the same problem, being two white box and two black box models. The complete description of these parameters is accessible in the official documentation [14], while we report a brief explanation below. The configurations we use are shown in Table 1; noteworthy, each individual attack has modifiable parameters to tweak in order to be more effective against the target model or have a more efficient computation of the adversarial example.

*Spatial Transformation:* The objective is to find the minimum spatial transformation that causes misclassification of the RGB image. The image is rotated by a $\theta$ angle and shifted of ($\delta u$, $\delta v$) pixels. The shift is calculated as a percentage of the image size. It is a black-box attack, needing only the class prediction for the input image.

*HopSkipJump:* It starts from a big perturbation and aims to reduce it to a minimum that still causes misclassification. The perturbation is reduced by iterations of binary searches. Each iteration produces a new perturbation, smaller than the previous one, and it stops when the boundary between the target class and the original class is reached. The distance from the original input is computed by using a norm. It is a black-box attack, needing only the class prediction for the image.

*NewtonFool:* It is a gradient-descent based algorithm that aims to find the perturbation that minimizes the probability of the original class. It is built under the assumption that, "nearby" the original data point, there is another point where the confidence probability in the "correct class" is significantly lower [13]. The tuning parameter $\eta$ determines how aggressively the gradient descent attempts to minimize the probability of the original class. It is a white-box attack, because the attacker requires the output of the softmax function which assigns decimal probabilities to each class in a multi-class problem.

*Basic Iterative Method.* It is an iterative version of the Fast Gradient Method (FGM, [15]), which produces an adversarial example by calculating the perturbation that maximizes the loss (with respect to the loss of the input image). The Basic Iterative Method extends the FGM attack by applying it multiple times with small step sizes. Each intermediate result is cropped to ensure that it stays within the limits established by a hyper parameter ε, that sets the amount of perturbation allowed in the target image. It is a white-box attack because it accesses the model to compute the loss at each step.

**Table 1.** Configurations used for each attack. A detailed description of the meaning of each parameter is available in [14].

| Attack | Configuration |
|---|---|
| Spatial Transformation | max shift = 80%, number of shifts = 1, max rotation = 160°, number of rotations = 1 |
| HopSkipJump | max iterations = 10, max_eval = 1000, init_eval = 100, init_size = 100, norm=2 |
| NewtonFool | max iterations = 10, $\eta$ = 0.01 |
| Basic Iterative Method | $\varepsilon$ = 0.2, $\varepsilon$_step = 0.1, max_iter = 20 |

Other relevant evasion attacks, that we left out for this work, are based on introducing modifications on the environment, rather than on the acquired images. An example is the Adversarial Patch [16] attack, which is based on crafting patches that can be applied on (or next to) objects, such that the trained agent is confused. This category of attacks was ultimately discarded, because they required a completely different approach that includes modifying assets of the simulations with the help of the Unreal Engine 4 editor.

### 3.2 Integration of ART in Learning by Cheating

To effectively use ART with LbC, we first define an attack module. This module contains two main functions: *load_model* and *load_attack*.

Function *load_model*(*model_path*) loads the weights of the pretrained agent. It takes a file path as input. This path must point to two files: a *config.json* and the PyTorch model *.pth* containing the model weights. Clearly, the PyTorch model is the trained agent LbC, and the *config.json* is created with the model parameters. The *load_model* function returns an ART *PyTorchClassifier*, a wrapper class that contains model weights. This wrapper class allows the interaction between the attacks and the trained agent developed in PyTorch; in other words, it allows executing the successive *load_attack* on the target model.

Function *load_attack*(*classifier*, *attack*) selects one of the attacks. It requires two parameters: an instance of an ART *PyTorchClassifier* (generated at the previous step) and a string which identifies the attack. The function returns an instance of the class associated with the attack.

The injection of the attacks is done by directly modifying the source code of LbC. The trained agent is implemented in the *ImageAgent* class of LbC [3], [30]. This class is modified by adding two extra fields: *self.adv* and *self.attack*. These fields contain the targeted model and the chosen attack, loaded by calling the two functions described before.

To execute the attacks, we have also to modify the main decision loop of LbC, that is used to generate the waypoints on the basis of the collected observations. More specifically, we modify the *run_step* method from Listing 1 that generates the waypoints. In fact, the *run_step* takes as input observations the RGB image, the speed of the vehicle and the supervisor command, and feed them to the neural network to generate the next waypoints. In our version, the RGB image is modified before being passed to the network. This is done by introducing two instructions inside the *run_step*:

```
_rgb = self.attack.generate(rgb.cpu())
_rgb = torch.FloatTensor(_rgb)
```

The first function creates the adversarial example; the implementation is different for each attack. The second instruction is necessary because the network processes float tensors while the *generate* function returns an array.

To better integrate ART with LbC, we need to slightly change the network decision function. In LbC this functionality is defined in the *forward* propagation function (the forward function defines how the model is going to be run, from input to output [18]) that is implemented in the *ImagePolicyModelSS* class. It creates the waypoints and it is also used by the *self.attack.generate* function to make the adversarial image. The *self.attack.generate* function only takes one parameter (the image) and it does not have access to speed and command values. This was preventing a proper invocation of the *forward* function, crashing the entire execution. The issue is solved changing how the *forward* function behaves: in the modified agent it takes only the image as a parameter, while speed and command are acquired through global variables.

## 4  Injection of Faults in a self-driving agent

We describe how we inject in the convolutional operations of the LbC trained agent.

### 4.1  Perturbation models

We describe our neuron and weight injections with PyTorchFI below. We used both neuron and weight injections in our experiments with different perturbation models: some of those are default models, others are customized.

*Neuron injection.* This perturbation changes the value of neurons insides the neural network with a value specified by the user. It can be on a single location or spread across the network. To perform neuron injections, the following parameters are required by PyTorchFI [24], [31]:
1. *conv_num*: the target convolutional layer.
2. *batch*: the batch in which the fault injector should run. This is set to 1 in our case, as the trained agent uses batches of a single input image.
3. *c*: the channel number.
4. *h:* the height of the input image, 160 pixels in our case.
5. *w*: the width of the input image, 384 pixels in our case.
6. *value:* the value(s) that the injector should set within the tensor.

PyTorchFi provides multiple perturbation models to simplify the injection process for the user. We use the *random_neuron_inj* and *random_inj_per_layer*. Function *random_neuron_inj* selects a single, random neuron in the network and changes its value to a random value between a range specified by the user (the default range is [-1, 1]). Instead, *random_inj_per_layer* selects a single, random neuron in each convolutional layer of the network, and changes its value to a random value between a range specified by the user (the default range is again [-1, 1]).

*Weight injection.* Weight injection has the same functionality of neuron injection, but on weights. The parameters required are the same as for neuron injection, aside the replacement of batch number with the kernel number.

Because of compatibility problems we couldn't use the utility function *random_weight_location,* which returns a tuple that represents the target location of the perturbation. Instead, we implement a custom method that calculates the injection location. We create an alternative function *random_weight* that returns a tuple containing the values that represent the location inside the network. The output of the function is a tuple including the following elements:

1. Convolutional layer number *conv_num*.
2. kernel number *k*: this is a random value between [0, 60]. We found this interval experimentally, through multiple simulations.
3. Channel number *c*, height *h* and weight *w* of the input image.

Finally, we defined a custom function for weight injection that is essentially a copy of the default PyTorchFI weight perturbation model, with the difference that it calls our random weight locator.

## 4.2 Application of PyTorchFi in LbC

We need to slightly modify the *ImagePolicyModelSS* class [30] from LbC, which is executed during the *run_step* of Listing 1, by adding the following: i) import *fault_injection* and the desired perturbation models from PyTorchFi; ii) create the PyTorchFi model *pfi_model*, which is essentially a copy of the Resnet34 backbone of LbC, where injections shall be performed; iii) perform the injection on *pfi_model*, for example simply invoking the PyTorchFI API instructions:

*inj = random_inj_per_layer(pfi_model, min_val = -10, max_val =10)*
*h = inj(image) # image is acquired from frontal camera*

## 5 Experiments and Results

### 5.1 Description of the experimental campaign

The experimental campaign is based on the *NoCrash* benchmark from [19], designed to test the ability of vehicles to handle complex events caused by changing traffic conditions (e.g., traffic lights) and dynamic agents in the scene. In multiple runs, a target vehicle must reach a destination position *B* from a starting position *A* before a timeout expires, and under different weather conditions. The timeout value is the time required to cover the distance from *A* to *B* at an average speed of 10 Km/h as in [1], [3]. For each individual run, the success criteria is that the destination *B* is reached before expiration of the timeout. The failure criteria is whenever the vehicle collides or the timeout expires. We include a modification of the *NoCrash* benchmark that halts the run whenever a collision occurs [4], because in our work we prioritize safety over travelled distance.

The benchmark implementation available at [30] records and saves videos of each run (for example, two screenshots are reported in Fig. 1), together with detailed logs.

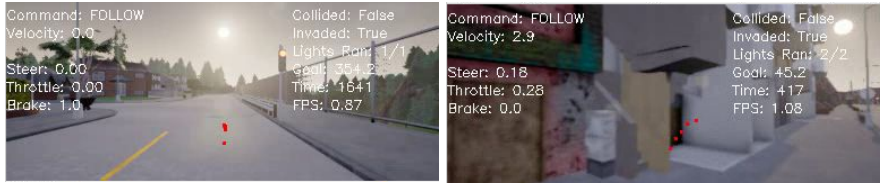

**Fig. 1.** Sample frame from the videos recorded. *Left*: the car is driving correctly on the intended lane. *Right*: an accident caused by the injection of HopSkip Jump. The vehicle does not steer as much as it should, and it goes out of the road, colliding with a building.

These videos were analyzed to visually confirm the results reported in the log and the effectiveness of the attacks and faults injected. To measure their effects, we measure the number of runs completed (the success criteria), the number of ignored red lights, and the number of collisions.

The target town we select are the Carla towns Town01 and Town02, that are basic town layouts with all "T" junctions. Town01 is the town used for training of the learned agent LbC [3], while Town02 is only used for testing. Scenarios Town01v1 and Town02v1 includes a single right turn, while scenarios Town01v2 and Town02v2 includes multiple turns and crossings (Table 2). We select different values for pedestrians and vehicles, and different weather conditions as reported in Table 2. Finally, we always use the same randomization seed so that spawning positions of vehicles and pedestrians are the same in the repeated runs.

The experiments are organized in two phases. In the first phase, we perform *golden runs* to produce clean data i.e., we execute the simulation runs without introducing any attack or fault. Results of the golden runs are in Table 2: in this case, most of the runs are successfully completed with a limited number of collisions, and only few red lights are ignored. Visually, we can confirm that the driving is stable: typically, the vehicle is well-placed in the middle of its lane. The vehicle only struggled in Town02-v2, where the adverse weather may have played a part in the increase of instability. In this case, most of the failures are caused by collisions, with a limited number of timeouts.

The second phase repeats the same runs of the previous phase, but with the injection of attack as described in Section 3, and of faults as described in Section 4. Attacks and

**Table 2.** Configuration details on the planned runs, and the results obtained for the golden runs, with 30 repetition of each configuration. The same configurations will be used for the attacks and faults injection campaigns.

|            |        |          |             |                         | Golden runs |           |
|------------|--------|----------|-------------|-------------------------|-------------|-----------|
| Run Name   | Town   | Vehicles | Pedestrians | Weather                 | Completed   | Collisions |
| Town01-v1  | Town01 | 20       | 50          | Clear Noon, Wet Noon,   | 27          | 2         |
| Town02-v1  | Town02 | 15       | 50          | Hard Rain, Clear Sunset | 26          | 2         |
| Town01-v2  | Town01 | 20       | 50          | Wet Sunset, Soft-Rain   | 28          | 2         |
| Town02-v2  | Town02 | 15       | 50          | Sunset                  | 15          | 13        |
|            |        |          |             | **Sum of runs**         | 96          | 19        |
|            |        |          |             | **Runs in timeouts**    | 5           |           |
|            |        |          |             | **Red lights fail**     | 28/417      |           |

410

faults are injected at each simulation step during the run. Results of the second phase are described in Section 5.2 and Section 5.3.

The simulations were executed on an Intel i9-9920X@3.50GHz CPU with Nvidia Quadro RTX 5000 GPU.

## 5.2 Adversarial attacks injection: results

An overview of the results for the injection of adversarial attacks is in Table 3, and discussed below.

*HopSkipJump (HSJ).* The designated route is generally followed properly by the vehicle. However, visually checking the videos, we can observe that the trajectory becomes much more unstable. The waypoints generated constantly change direction and this causes major issues, especially in curves. As for example in Fig. 1, it may happen that the vehicle does not steer as much as it should, going off-road and leading to collisions. Most of the crashes happened with the bad weather. The instability of the waypoints also leads to an increase of ignored red lights, with the vehicle entering crossroads occupied by other vehicles. Only two runs ended by timeout.

*Spatial Transformation (STA).* The results are comparable to the golden runs. Few red lights were ignored, and the collisions happened in similar spots. In some runs the waypoint were slightly different but the trajectory was very stable. Five runs were not completed because of a timeout.

*Basic Iterative Method (BIM).* The driving is seriously compromised. In each run, it immediately starts turning right, going off-road and colliding with the surroundings. No semaphore was even reached, and no run was completed.

*NewtonFool (NF).* The agent's driving is unreliable. A very low stability in the trajectory causes serious problems in staying inside the lane. This caused a significant increase in collisions and in red lights ignored. Only one run ended with a timeout.

The results show a general decrease in reliability of the car when it is injected with the attacks. We can see the two extreme cases in the injection of ST and BIM. While the first seems to not cause any problem to the agent, the latter is very detrimental. During the other two attacks, instead, the car still functions on a basic level, but becomes much more unstable with respect to the golden runs, especially in turns and with bad weather. This indicates that these kinds of attacks can pose a real threat to the safety of autonomous vehicles and needs to be considered during development.

**Table 3.** Overview of results for the 4 injected attacks; each configuration was repeated 30 times for each attack.

|  | HSJ | | STA | | BIM | | NF | |
| --- | --- | --- | --- | --- | --- | --- | --- | --- |
| **Run Name** | Completed | Collisions | Completed | Collisions | Completed | Collisions | Completed | Collisions |
| Town01-v1 | 27 | 3 | 28 | 1 | 0 | 30 | 13 | 17 |
| Town02-v1 | 15 | 14 | 26 | 2 | 0 | 30 | 7 | 22 |
| Town01-v2 | 16 | 13 | 26 | 4 | 0 | 30 | 0 | 30 |
| Town02-v2 | 4 | 26 | 17 | 11 | 0 | 30 | 0 | 30 |
| **Runs sum** | **62** | **56** | **97** | **18** | **0** | **120** | **20** | **99** |
| **Timeouts** | 2 | | 5 | | 0 | | 1 | |
| **Red lights fail** | 162/339 | | 26/420 | | 0 | | 156/227 | |

### 5.3 Faults injection: results

An overview of the results for the injection of faults using PyTorchFI is in Table 4 to Table 6, and discussed below.

*Random neuron injection.* The more we increase the range of values, the more the vehicle becomes unstable. This can be easily observed comparing the left and right side of Table 4. On the right side of Table 4, with a range of [-10000, 10000] we can see that not only the self-driving agent fails to reach the goal, but all the simulations terminate almost immediately. This can be also verified by looking at the number of traffic lights encountered.

*Random neuron injection per layer.* It's expected that this injection causes more problems, because it injects a fault on each layer of the neural network. This is true even if the numerical value injected is much lower than in the previous case. As confirmed in Table 5 (right side), within the range of [-100, 100] we already obtained similar results as the single location injection with range of [-10000, 10000] (Table 4, right side).

*Random single weight injection.* The effects of weight injection are less visible than neuron injection. This can be observed in Table 6, where we have tried to increase the range of the injected erroneous value from [-1000, 1000] to [-10000, 10000], but without observing a significant change in the vehicle behavior: in fact, even though we can visually verify that the trajectory is unstable, the success rate is close to the golden runs in both cases.

**Table 4.** (*Left*) Single random location neuron injection with values between [-1000, 1000]. (*Right*) Single random location neuron injection with values between [-10000, 10000].

| Town | Success rate | Collision | Ignored lights | Town | Success rate | Collision | Ignored lights |
|---|---|---|---|---|---|---|---|
| Town01-v3 | 4/5 | 1/5 | 5/11 | Town01-v3 | 0/5 | 5/5 | 0/0 |
| Town01-v4 | 0/5 | 5/5 | 1/6 | Town01-v4 | 0/5 | 5/5 | 0/0 |
| Town02-v3 | 0/5 | 5/5 | 3/11 | Town02-v3 | 0/5 | 5/5 | 4/6 |
| Town02-v4 | 0/5 | 5/5 | 1/9 | Town02-v4 | 0/5 | 5/5 | 2/4 |

**Table 5.** (*Left*) Random neuron injection per layer with values between [-50, 50]. (*Right*) Random neuron injection per layer with values between [-100, 100].

| Town | Success rate | Collision | Ignored lights | Town | Success rate | Collision | Ignored lights |
|---|---|---|---|---|---|---|---|
| Town01-v3 | 4/5 | 1/5 | 7/11 | Town01-v3 | 0/5 | 5/5 | 2/6 |
| Town01-v4 | 1/5 | 4/5 | 4/7 | Town01-v4 | 0/5 | 5/5 | 6/6 |
| Town02-v3 | 0/5 | 4/5 | 6/13 | Town02-v3 | 0/5 | 5/5 | 4/5 |
| Town02-v4 | 0/5 | 5/5 | 4/5 | Town02-v4 | 0/5 | 5/5 | 3/5 |

**Table 6.** (*Left*) Random weight injection with values between [-1000, 1000]. (*Right*) Random weight injection with values between [-10 000, 10 000].

| Town | Success rate | Collision | Ignored lights | Town | Success rate | Collision | Ignored lights |
|---|---|---|---|---|---|---|---|
| Town01-v3 | 5/5 | 0/5 | 0/11 | Town01-v3 | 5/5 | 0/5 | 0/11 |
| Town01-v4 | 5/5 | 0/5 | 0/11 | Town01-v4 | 4/5 | 1/5 | 1/11 |
| Town02-v3 | 4/5 | 0/5 | 3/28 | Town02-v3 | 4/5 | 1/5 | 2/27 |
| Town02-v4 | 2/5 | 3/5 | 0/23 | Town02-v4 | 3/5 | 2/5 | 1/23 |

In our results, the neuron injection per layer has the highest impact, followed by the single random location neuron injection, while the most "ineffective" model is the single random weight injection. This is most likely because the number of weights in a neural network is immensely big, and not all of them are used to process inputs.

## 6    Limitations and impact for real-world scenarios

We discuss two potential limitations of our study, also at the light of related works and of the possible impact for real-world scenarios.

The first limitation concerns the *amount of simulations and the selected configurations*. Input parameters to the faults and attacks injector are essential, and different configurations would clearly lead to significantly different results. This is true for both attacks and faults injection. The parameters of the attacks in Table 1, and the configuration for the generation of faults in Section 5, can be significantly varied and consequently can lead to i) different alterations to the input image and ii) different decisions of the trained agent. Our experimental campaign is clearly not inclusive of all the possible configurations for faults and attacks. Nonetheless, the validity of the initial objectives of our study are still valid, which are i) give evidence that a self-driving agent can fail due to software faults affecting the trained agent and due to (classifier-oriented) adversarial attacks, and ii) show an approach to quickly study these faults and attacks in a simulated environment.

The second limitation is on the *feasibility of adversarial attacks and the fault mode* in the considered scenarios. Concerning adversarial attacks, while other authors have very recently hypothesized their criticalities for autonomous driving [7], [5], [9], it is evident that implementing these attacks requires a high level of control on the attacked system. Performing an adversarial attack on a vehicle would require at least to capture the images provided by a camera and alter them before they are further processed by the trained agent. As communication between sensors and processing units for safety-critical components in vehicle is typically cabled, this would require a severe physical hacking on the vehicle, and consequently adversarial attacks are presently very difficult to apply (a different aspects is for Adversarial Patches [16], which instead can be located along the road as demonstrated in [22]). However, as in the recent history we have witnessed also remote hacking of vehicles [21], such attack surface should not be neglected a-priori, and we believe our work can contribute as a warning in this direction.

Failures instead may be the consequence of bugs in neural network software [26], [27] or accelerators faults (GPUs) [23], [25]. Several recent works have raised a warning about their dangerousness [29], overall describing the possible occurrence of software and hardware faults in a similar fashion as for other safety-critical hardware and software parts of the system. Consequently, we do not need to further motivate about the possible occurrence of faults. However, the exploration of fault modes and the relative consequences on trained agents, and in particular self-driving agents, still requires investigation, as demonstrated by the many recent works on the subject e.g., [27], [28]. Again, our work can act as i) a warning on the effect in safety-critical systems, as we show that even one persistent fault injected in just one of the network layers can jeopardize functional safety, and ii) an approach to study the impact of such faults in trained agent, relying on simulated environments.

## 7  Conclusions

This paper describes our experience with the injection of attacks and software faults in a self-driving agent running on the Carla simulator, entirely relying on open source tools. With respect to "traditional" software, self-driving agents have peculiarities such that they exposes new attack surfaces and fault modes. The objective of this experience report is to show how these attack surfaces and new fault modes can be exploited, making it explicit that this can violate vehicle safety.

While results are clearly restricted to the set of configurations used and to the self-driving agent in use, they still clearly show the possible detrimental effects of adversarial attacks and faults, as well as the need to protect and test autonomous systems against them.

Finally, the paper aims to show how self-driving agent can be tested, relying on existing tools and a controlled environment, through a carefully described experimental campaign. Overall, our injection activities required just few modifications to the trained agent (mostly related to the forward function), and minimal knowledge on the underlying simulator.

**Acknowledgment.** This work has been partially supported by the project POR-CREO SPACE "Smart PAssenger CEnter" funded by the Tuscany Region.